\documentclass{article}
\usepackage{spconf,amsmath,graphicx}
\usepackage{enumerate}
\usepackage{indentfirst} 
\usepackage{amssymb}
\usepackage{bm}
\usepackage{multirow}
\usepackage{color,xcolor}
\newcommand{\Diff}[0]{{\rm Diff}}
\usepackage{algorithm}  
\usepackage[noend]{algpseudocode}
\usepackage{algorithmicx,algorithm}

\usepackage{algpseudocode}  
\usepackage{tikz,xcolor,hyperref}
\usepackage{color}

\usepackage{enumitem}
\setlist{nosep, leftmargin=14pt}

\usepackage{mwe} 
\usepackage{framed} 

\title{Hybrid Atlas Building with Deep Registration Priors}
\name{Nian Wu$^{\star}$ \quad Jian Wang$^{{\dagger}}$ \quad Miaomiao Zhang$^{{\dagger}{\ddagger}}$ \quad Guixu Zhang$^{{\star}}$ \quad Yaxin Peng$^{\ast}$  \quad Chaomin Shen$^{{\star}}$}

\address{$^{{\star}}$Department of Computer Science and Technology, East China Normal University, China \\
    $^{\dagger}$Department of Computer Science, University of Virginia, USA\\
    $^{\ddagger}$Department of Electrical and Computer Engineering, University of Virginia, USA\\
$^{\ast}$Department of Mathematics, College of Science, Shanghai University, China}

\begin{document}
%
\maketitle
\begin{abstract}
Registration-based atlas building often poses computational challenges in high-dimensional image spaces. In this paper, we introduce a novel hybrid atlas building algorithm that fast estimates atlas from large-scale image datasets with much reduced computational cost. In contrast to previous approaches that iteratively perform registration tasks between an estimated atlas and individual images, we propose to use learned priors of registration from pre-trained neural networks. This newly developed hybrid framework features several advantages of (i) providing an efficient way of atlas building without losing the quality of results, and (ii) offering flexibility in utilizing a wide variety of deep learning based registration methods. We demonstrate the effectiveness of this proposed model on 3D brain magnetic resonance imaging (MRI) scans.
\end{abstract}
\section{Introduction}
\label{sec:intro}
Image atlas (also known as mean template) has critical values in medical applications, as it provides an unbiased coordinate system for template-based image segmentation~\cite{wang2005construction,gao2016image}, statistical analysis of anatomical shape variability~\cite{zhang2015bayesian,zhang2016low}, and population studies of brain development or neurodegenerative disease progression~\cite{YING2014626,hong2017fast}.
In its simplest form, an atlas was randomly selected from a group of images with intrinsically biased anatomy towards a single subject~\cite{FLETCHER2007250, VERCAUTEREN2009S61}. To address this issue, unbiased atlas building methods have been developed to estimate an averaged image over the entire datasets~\cite{joshi2004unbiased,FLETCHER2009S143,zhang2013bayesian}. The problem of unbiased atlas building is formulated as a minimization of image registration between the estimated atlas and individual subjects~\cite{joshi2004unbiased}. While the aforementioned methods have undergone technical developments with rigorous theory support and mathematical guarantees, their practical applicability is severely limited (particularly in large-scale and high-dimensional image spaces). The iterative gradient search of registration between an estimated atlas and each individual image on dense coordinates makes the computation challenging.   

Despite fast registration algorithms such as DARTEL~\cite{ashburner2007fast} and FLASH~\cite{zhang2019fast} were developed to alleviate this problem, the atlas building process still takes tens of minutes or even longer to converge. Recent advances in deep learning provide another research line of atlas building by simultaneously learning the atlas and registration parameters from deep networks~\cite{dalca2019learning,hinkle2018diffeomorphic}. However, these methods require to re-train the entire network when a group of new testing images arrives.

In this paper, we present a hybrid atlas building model that leverages learned prior information from pre-trained deep networks of image registration. To the best of our knowledge, we are the first to introduce a predictive framework of atlas building with deep neural network priors. In particular, we develop an integrated framework that efficiently merges model-based atlas learning with fast predictive registration networks trained on a large collection of images. In contrast to existing approaches of estimating atlas, our newly proposed model has a set of advantages, i.e., 
\begin{enumerate}[label=(\roman*).]
\item It reduces the computational complexity of atlas building by eliminating the computational bottleneck of gradient-based optimization of registration algorithms.
\item It is generic to a wide variety of deep learning based registration networks, for example, Quicksilver~\cite{YANG2017378}, DeepFLASH~\cite{Wang_2020_CVPR}, and VoxelMorph~\cite{10.1007/978-3-030-00928-1_82}.
\item It can be easily generalized to a mixture or hierarchical modeling of atlas building for advanced image analysis tasks, i.e., multi-atlas building~\cite{zhang2015mixture} and population-based longitudinal image studies~\cite{singh2013hierarchical}. 
\end{enumerate}
To demonstrate the effectiveness of our proposed model, we run experiments on 3D real brain MRIs with a wide variety of deep registration networks. We also compare the model's performance with a recent model-based fast atlas building method~\cite{zhang2019fast} and deep learning based atlas building~\cite{dalca2019learning,hinkle2018diffeomorphic}.

\section{Background: Deformable Image Registration}
\label{sec:background}
We focus on diffeomorphic image registration, as it provides a one-to-one smooth and invertible smooth mapping (a.k.a. diffeomorphism) between images~\cite{beg2005computing}. Diffeomorphic transformations are desirable in many medical applications to preserve the topological structure of studied subjects.

Given a source image $S$ and a target image $T$ defined on a $d$-dimensional torus domain $\Omega = \mathbb{R}^d / \mathbb{Z}^d$ ($S(x), T(x):\Omega \rightarrow \mathbb{R}$). The problem of diffeomorphic image registration is to find the shortest path, i.e., geodesic, to generate time-varying diffeomorphisms $\{\phi_t(x)\}: t \in [0,1] $, such that $S \circ \phi^{-1}_1$ is similar to $T$, where $\circ$ is a composition operator that performs interpolation on $S$ by the smooth mapping $\phi^{-1}_1$.  This can be solved by minimizing an explicit energy functional over the transformation fields $\phi_t$ as
\begin{eqnarray}
\label{eq:tenergy}
E(\phi_t) = \frac{1}{2\sigma^{2}}\, \text{Dist}(S \circ \phi_1^{-1}, T) + \text{Reg}(\phi_t).
\end{eqnarray}
Here Dist(·,·) is a  distance function that measures the dissimilarity between images, Reg($\cdot$) is a regularization term that enforces the smoothness of transformation fields, and $\sigma$ is a positive weighting parameter. Widely used distance functions include the sum-of-squared intensity differences ($L_2$-norm)~\cite{beg2005computing}, normalized cross correlation (NCC)~\cite{avants2008symmetric}, and mutual information (MI)~\cite{wells1996multi}. In this paper, we will use the sum-of-squared intensity differences. 

{\bf Time-dependent velocity fields.} In the setting of LDDMM (large diffeomorphic deformation metric mapping) framework~\cite{beg2005computing}, a path of diffeomorphisms $\{\phi_t\}$ is generated by a flow of time-dependent velocity fields $\{v_t\}$ in the tangent space of diffeomorphism denoted as $V = T\Diff(\Omega)$, i.e., 
\begin{eqnarray}
\label{eq:velocity}
\frac{d\phi_t}{dt} = v_t \circ \phi_t,  \,\, \text{s.t.}  \,\, \,\, \phi_0 = x.
\end{eqnarray}
The geodesic shooting algorithm~\cite{Vialard2011DiffeomorphicAE} states that a geodesic path of the diffeomorphic transformations can be uniquely determined by integrating the Euler-Poincaré differential equation (EPDiff)~\cite{arnold1966,miller2006} with an initial condition $v_0$
\begin{align}\label{eq:epdiff}
\frac{d v_t}{dt} = -K \left[\left( D v_t\right)^\mathrm{T} m_t + D m_t \, v_t + m_t \operatorname{div} v_t \right],
\end{align}
where $D$ denotes a Jacobian matrix and $\rm{div}$ is the divergence operator.  Here $K$ is the inverse operator of $L: V \rightarrow V^*$, which is a symmetric, positive-definite differential operator that maps a tangent vector $v \in V$ into the dual space $m \in V^*$.  

{\bf Stationary velocity fields.}
Another way to compute diffeomorphism is using stationary velocity fields~\cite{hernandez2009registration}. The transformation $\phi_t$ is parameterized by a stationary velocity field $w$,
\begin{eqnarray}
\label{eq:velocity_stationary}
\frac{d\phi_t}{dt} = w(\phi_t),   \,\, \text{s.t.}  \,\, \,\, \phi_0 = x.
\end{eqnarray}
The solution of Eq.~\eqref{eq:velocity_stationary} is identified with a group exponential map. More details are included in~\cite{hernandez2009registration}.

By parameterizing the transformation $\phi_t$ with an initial velocity field $v_0$, we can equivalently rewrite the optimization of Eq.~\eqref{eq:tenergy} as
\begin{eqnarray*}
E(v_0) = \frac{1}{2\sigma^{2}}\, \| S \circ \phi_1^{-1} - T \|_{L^2}^2 + \text{Reg}(v_0), \, \, \text{s.t. Eq.}~\eqref{eq:velocity} \, \text{or} \, \eqref{eq:velocity_stationary}.
\end{eqnarray*}
For a simplified notation, we will drop the time index in following sections, i.e., $v_0 \overset{\Delta}{=} v$.  

\section{Our Method: Hybrid Atlas Building with Deep Registration Priors}
\label{sec:pagestyle}
In this section, we introduce a hybrid model of atlas building that incorporates learned regularization priors from pre-trained deep networks. We show that our model can be implemented using independent registration modules. Therefore, changing the prior model only involves the implementation of image registration. That is to say, our framework can be used to match a wide variety of priors with any suitable registration model. 

Consider a group of images $\{I_1, \dots, I_N\}$, the problem of atlas building is to find a mean image $\hat{I}$ (known as Fr{\'e}chet mean~\cite{joshi2004unbiased}) that requires a minimal amount of energy to map each subject $I_i$, $i \in \{1, \cdots, N\}$ to the atlas space. With a pre-learned registration prior $g_\theta(\cdot, \cdot)$ (where $\theta$ represents network parameters), we formulate a joint objective function to integrate atlas building with predictive image registration as 
\begin{align}\label{eq:atlas_energy_n}
E(\hat{I}, v_i) = & \frac{1}{N}\,\sum_{i=1}^{N} \frac{1}{2\sigma^{2}}\, \| \hat{I} \circ \phi_i^{-1} - I_i \|_{L^2}^2 + \frac{1}{2} \| v_i \|_V^2  \nonumber \\
& + \frac{1}{2} \lambda \| v_i - g_\theta(\hat{I}, I_i)\|_V^2, \, \, \text{s.t. Eq.}~\eqref{eq:velocity} \, \text{or} \, \eqref{eq:velocity_stationary}, 
\end{align}
where $\lambda$ is a parameter controlling the fidelity of each velocity and learned registration priors. The $\| \cdot \|_V$ represents a Sobolev space that enforces smoothness of the velocity fields. A common formulation is that $\| v_i \|^2_V = \langle L v_i, L v_i \rangle$, with $L$ being a differential operator such as Gaussian or Laplacian.    

In order to solve the problem~\eqref{eq:atlas_energy_n} efficiently, we adopt an alternating minimization approach~\cite{nocedal2006numerical}, where $v_i$ is first minimized for a fixed atlas image $\hat{I}$ and vice versa, as follows
\begin{align}
\label{eq:optimization}
   v_i^j &= \text{argmin}_{v_i} \, E(\hat{I}^{j-1}, v_i), \nonumber  \\
   \hat{I}^j &= \text{argmin}_{\hat{I}} \, E(v_i^j, \hat{I}), 
\end{align}
where $j$ denotes the $j$th iteration. 

\subsection{Inference} 
We employ a gradient decent algorithm to minimize the subproblems of Eq.~\eqref{eq:optimization}. The two main steps of our inference are described below. 
\begin{enumerate}[label=(\roman*).]
\item {\bf Closed-form update for registration.} By setting the gradient terms with respect to $v_i$ to zero, we obtain a simple closed-form update for image registration based on the pre-learned priors as
\begin{equation} \label{eq:vupdate}
 v_i^j = \frac{\lambda}{1+\lambda} g_\theta(\hat{I}^{j-1}, I_i).   
\end{equation}
\item {\bf Closed-form update for atlas.} By setting the gradient terms with respect to $\hat{I}$ to zero, we have a closed-form update for atlas as
\begin{equation}\label{eq:atlasupdate}
\hat{I}^j = \frac{\sum^N_{i=1}I_{i} \circ \phi_i^j|D\phi_i^j|}{\sum^N_{i=1}|D\phi_i^j|},
\end{equation}
where $|\cdot|$ denotes a determinant operator.
\end{enumerate}

\begin{figure*}[!t]
\centering
\includegraphics[width=1.0\textwidth]{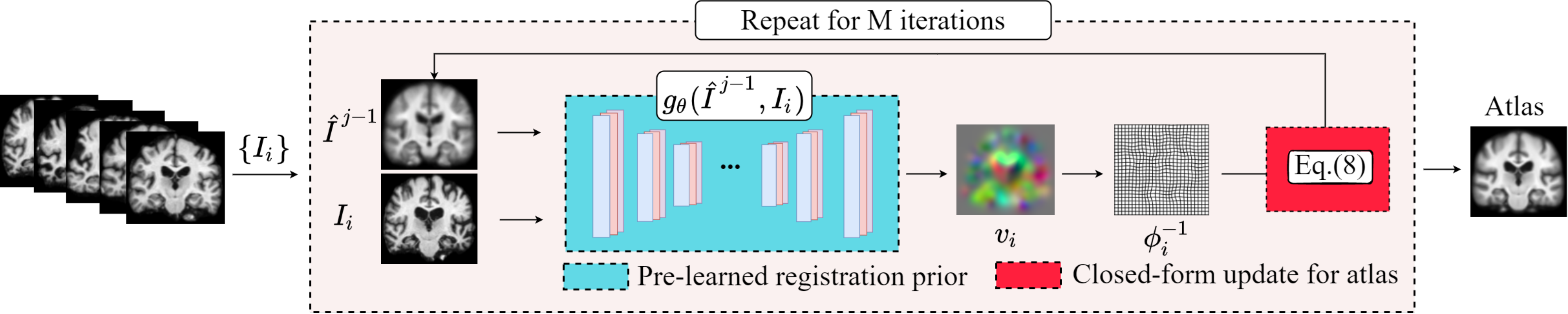}\par
\caption{Visualization of our hybrid atlas building model with deep registration priors.} 
\label{fig1}
\end{figure*}

We visualize the proposed hybrid atlas building model in Fig.~\ref{fig1}. A summary of our derived inference algorithm is presented in Algorithm ~\ref{alg:hab}.

\begin{algorithm}[!h]
  \caption{Hybrid Atlas Building.}
  \label{alg:hab}  
  \begin{algorithmic}[1]  
    \Require
      {A group of $N$ images $\{ I_i \}_{i=1}^{N}$.}
    \Ensure  
      {The optimal atlas $\hat{I}$.}  
    \State Initialization: averaging the $N$ inputs as initial atlas $\hat{I}$.
    \Repeat
          \For{$i=1 : N$}
            \State 
            $ v_i^j = \frac{\lambda}{1+\lambda} g_\theta(\hat{I}^{j-1}, I_i);$
          \EndFor
          \State 
          $\hat{I}^j = \frac{\sum^N_{i=1}I_{i} \circ \phi_i^j|D\phi_i^j|}{\sum^N_{i=1}|D\phi_i^j|};$
    \Until     
    \Return $\hat{I}$.
  \end{algorithmic}  
\end{algorithm}

\section{Experimental Evaluation}
We demonstrate our model on real 3D brain MR images. We first employ three the-state-of-the-art deep registration networks Quicksilver~\cite{YANG2017378} (HAB-QS), VoxelMorph~\cite{dalca2019learning} (HAB-VM), and DeepFLASH~\cite{Wang_2020_CVPR} (HAB-DF) as the prior models. We then compare their performances with a model-based atlas building algorithm FLASH~\cite{zhang2019fast} and two other deep learning based atlas building methods
VM-Atlas (the unconditional version)~\cite{dalca2019learning} and LagoMorph~\cite{hinkle2018diffeomorphic}. All estimates of atlases and time consumption are reported.
To quantitatively evaluate the quality of estimated atlases, we calculate the average NCC coefficient between the warped atlases and randomly selected $30$ testing images. 


{\bf Data.} We include $230$ T1-weighted 3D brain MRI scans (aged from $60$ to $90$) from Open Access Series of Imaging Studies (OASIS)~\cite{Fotenos1032} in our experiments. All MRIs were down-sampled to the size of $128^3$ with an isotropic resolution of $1.25 mm^3$. The scans have undergone skull-stripping, intensity normalization, bias field correction, and affine alignment. The dataset is split into $130$ training volumes for registration networks and the rest $100$ volumes for atlas building.

{\bf Registration training.} We randomly draw $1,740$ pairs of images from the training data to train Quicksilver (batch size as $1$, learning rate $\eta = 1e^{-4}$, and $200$ training epochs), VoxelMorph (batch the size as $1$, learning rate $\eta = 4e^{-4}$, and $200$ training epochs), and DeepFLASH (batch size as $1$, learning rate $\eta = 3e^{-4}$, and $300$ training epochs). All results are generated on Nvidia GeForce GTX 1080Ti GPUs.

\subsection{Results}
\begin{figure}[!htb]
\begin{minipage}[b]{1.0\linewidth}
  \centering
  \centerline{\includegraphics[width=8.5cm]{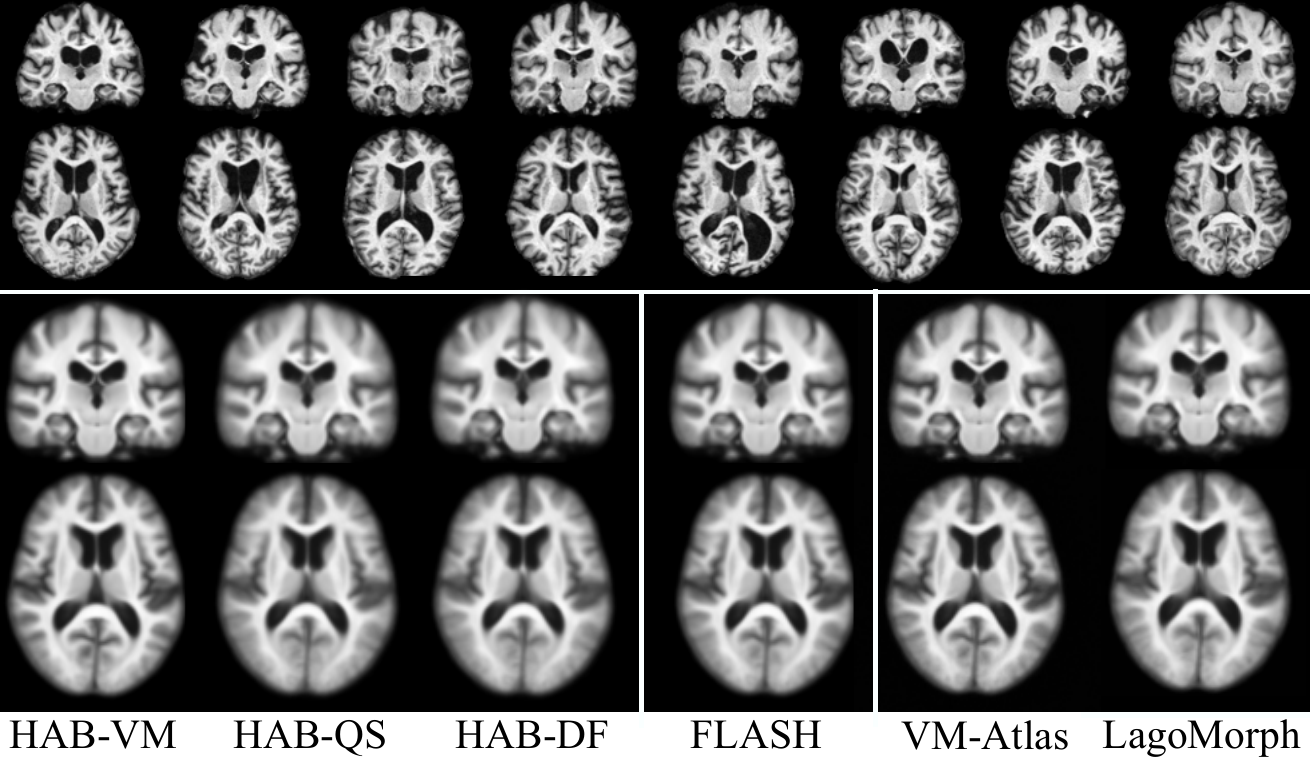}}
\end{minipage}
\caption{Top: examples of coronal and axial view of the 3D brain MRIs. Bottom: final atlases estimated by our hybrid model (HAB-VM, HAB-QS, HAB-DF), FLASH, and learning-based atlas building VM-Atlas and LagoMorph.}
\label{fig3d_all}
\end{figure}

Fig.~\ref{fig3d_all} visualizes the 3D brain atlases estimated by our model (HAB-QS, HAB-VM, and HAB-DF) and the baseline methods (FLASH~\cite{zhang2019fast}, VM-Atlas~\cite{dalca2019learning} and LagoMorph~\cite{hinkle2018diffeomorphic}). 
For our model, the NCC coefficients are ${0.92}$, ${0.92}$, and ${0.90}$ ({HAB-VM}, {HAB-QS}, and {HAB-DF}). For baseline methods, the NCC coefficients are $0.92$, $0.91$, and $0.93$ (FLASH, VM-atlas, and LagoMorph). It shows that our method is able to produce a comparable quality of atlas as baseline algorithms.

\begin{figure}[!htb]
\begin{minipage}[b]{1.0\linewidth}
  \centering
  \centerline{\includegraphics[width=8.5cm]{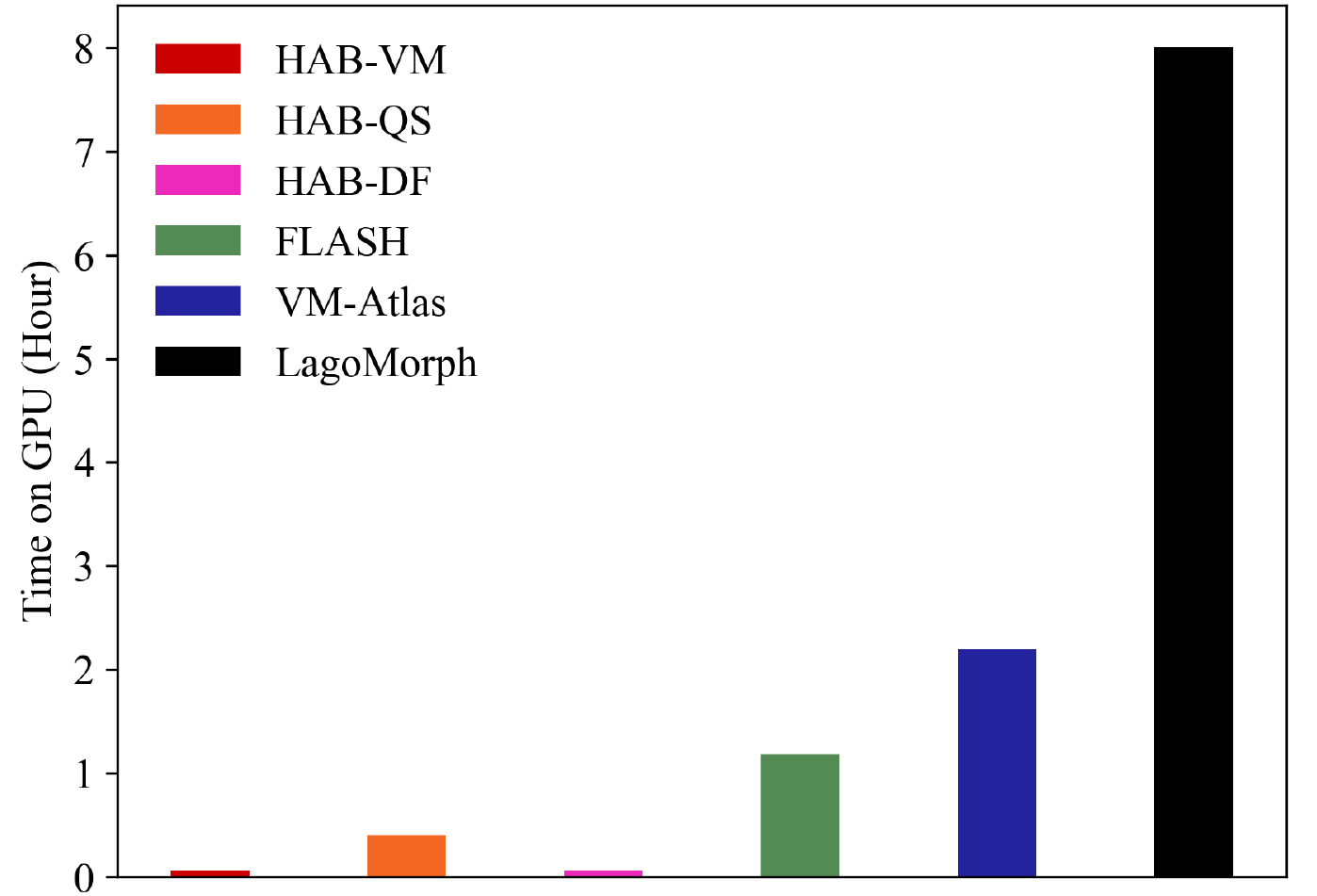}}
\end{minipage}
\caption{Time consumption of atlas building for all methods.}
\label{time}
\end{figure}

Fig.~\ref{time} provides the time consumption of atlas building on GPU servers. It shows that our proposed method dramatically reduces the time consumption by orders of magnitude compared to the baseline algorithms (either a fast version of conventional image registration FLASH, or the current learning-based approaches VM-Atlas~\cite{dalca2019learning} and LagoMorph~\cite{hinkle2018diffeomorphic}). It is worthy to mention that the state-of-the-art learning-based atlas building approaches~\cite{dalca2019learning,hinkle2018diffeomorphic} do not provide direct prediction to atlas. The networks need to be retrained when a new group of images arrives.

\begin{figure}[!htb]
\begin{minipage}[b]{1.0\linewidth}
  \centering
\centerline{\includegraphics[width=8.5cm]{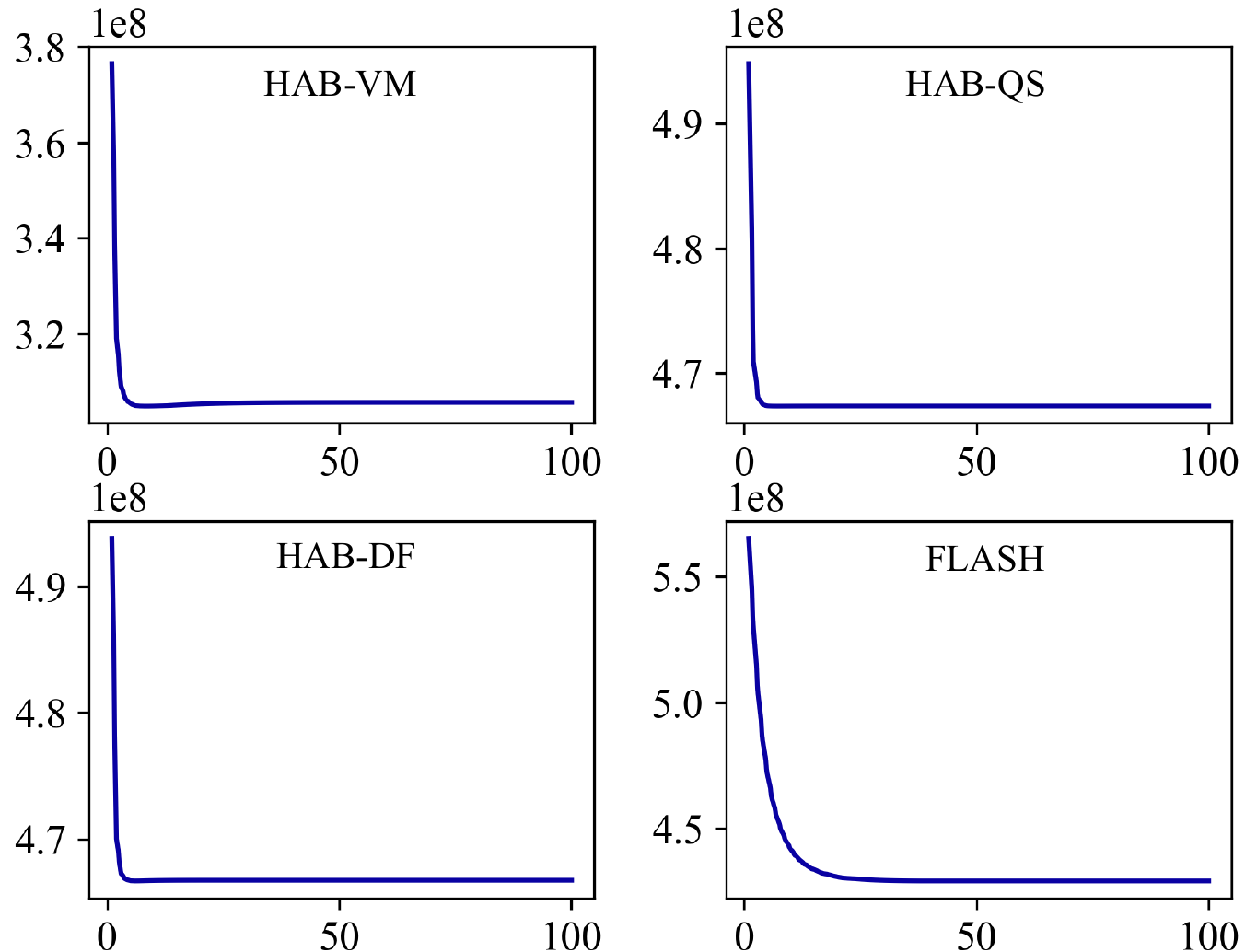}}
\end{minipage}
\caption{Convergence of atlas building by FLASH and our proposed model.}
\label{3d_energy}
\end{figure}

Fig.~\ref{3d_energy} displays the convergence graph of total energy on our proposed hybrid atlas building model. The comparison between our model and the conventional method FLASH indicates that our method achieves a comparable quality of atlas with a stable algorithmic convergence guaranteed.



\section{Conclusion}
\label{sec:Conclusion}
We present an efficient and hybrid atlas building model by integrating model-based learning with deep learning approaches. In contrast to traditional learning-based atlas building methods that perform an iterative numerical optimization to search for the registration results between the estimated atlas and individual images, we directly use the registration results predicted by a pre-trained deep neural network. It is worth mentioning our work aims to accelerate the time-consuming of atlas building process by well-trained deep registration priors while maintaining the high quality of atlas. The theoretical tool developed in our work is flexible to a wide variety of state-of-the-art registration networks. This work opens up the possibility for further integration of learning-based methods with more efficient deep neural networks, e.g., equipping the mixture or hierarchical modeling of atlas building with deep neural networks.
Another interesting future direction would be conducting statistical analysis of population-based image studies using our estimated results of atlas and individual deformations, i.e., for comparing group differences, or analyzing anatomical shape variability. 




\section{Acknowledgments}
This work was supported by National Science Foundation of China (61731009).

\bibliographystyle{IEEEbib}
\bibliography{TT_ISBIV2}

\end{document}